# Generic Preferences over Subsets of Structured Objects

**Maxim Binshtok**                                    MAXIMBI@CS.BGU.AC.IL
**Ronen I. Brafman**                                  BRAFMAN@CS.BGU.AC.IL
*Computer Science Department*
*Ben-Gurion University, Israel*

**Carmel Domshlak**                                DCARMEL@IE.TECHNION.AC.IL
*Faculty of Industrial Engineering and Management*
*Technion, Israel*

**Solomon E. Shimony**                                SHIMONY@CS.BGU.AC.IL
*Computer Science Department*
*Ben-Gurion University, Israel*

## Abstract

Various tasks in decision making and decision support systems require selecting a preferred subset of a given set of items. Here we focus on problems where the individual items are described using a set of characterizing attributes, and a generic preference specification is required, that is, a specification that can work with an arbitrary set of items. For example, preferences over the content of an online newspaper should have this form: At each viewing, the newspaper contains a subset of the set of articles currently available. Our preference specification over this subset should be provided offline, but we should be able to use it to select a subset of any currently available set of articles, e.g., based on their tags. We present a general approach for lifting formalisms for specifying preferences over objects with multiple attributes into ones that specify preferences over subsets of such objects. We also show how we can compute an optimal subset given such a specification in a relatively efficient manner. We provide an empirical evaluation of the approach as well as some worst-case complexity results.

## 1. Introduction

Work on reasoning with preferences focuses mostly on the task of recognizing preferred elements within a given set. However, another problem of interest is that of selecting an optimal *subset* of elements. Optimal subset selection is an important problem with many applications: the choice of feature subsets in machine learning, selection of a preferred bundle of goods (as in, e.g., a home entertainment system), finding the best set of items to display on the user's screen, selecting the best set of articles for a newspaper or the best members for a committee, etc.

Earlier work on this problem has mostly focused on the question of how one can construct an ordering over subsets of elements given an ordering over the elements of the set (Barberà, Bossert, & Pattanaik, 2004). The main distinction made has been between sets of items that are mutually exclusive, in the sense that only one can eventually materialize, and sets in which the items will jointly materialize. Our formalism is agnostic on this issue, although we are clearly motivated by the latter case. As Barberà et al. note, most past work focused on the case of mutually exclusive elements. This, for example, would be the case if we are selecting a set of alternatives from which some decision-maker (or nature) will ultimately choose only one (e.g., courses of action). However, there is a substantial body of work on the latter setting in which items might materialize jointly, and individual items are preferentially comparable.

This paper focuses on a somewhat different context for set-preference specification. First, we assume that the items from which our subsets are composed are structured, in the sense that some set of attributes is associated with them. For example, if the items are movies, these attributes could be the genre, language, year, director; if the "items" are politicians, the attributes could be the political views of the politicians on various topics, their party affiliation, their level of experience. Second, we require a generic preference specification, in the sense that it can be used with diverse collections of items. For example, if we are specifying guidelines for the composition of some committee, these guidelines are generic, and can be used to induce a preference relation over subsets of any given set of politicians, provided that the set of attributes is fixed. Third, we do not assume any preferential ordering over the individual items, although that can certainly be captured by one of the attributes describing the items.

An instructive example of the type of domain we have in mind is that of personalized online newspapers. First, the problem of selection for a newspaper is one of subset selection – we have to select a subset of the set of available articles to place in the newspaper. Second, the database of articles is constantly changing. Therefore, an approach that requires explicitly specifying preferences for the inclusion of each specific item is inappropriate, both because the number of such items is very large, and because this would require us to constantly change the preference specification as the set of items changes. Finally, we would not want to base our approach on a method for transforming an ordering over items into an ordering over subsets of items, because we do not want to have to rank each item, and because there are obvious instances of complementarity and substitutability. For instance, even if I prefer articles on Britney Spears to articles on any other topic, two very similar articles about her may be less interesting than a set comprising one about her and one about the Spice Girls.[1]

One recent work that considers a similar setting is that of desJardins and Wagstaff (2005), which works by specifying preferences over more abstract properties of sets. In particular, desJardins and Wagstaff offer a formalism for preference specification in which users can specify their preferences about the set of values each attribute attains within the selected set of items. One could assert whether the values attained by an attribute on the selected subset should be diverse or concentrated around some specific value. In addition, desJardins and Wagstaff also suggest a heuristic search algorithm for finding good, though not necessarily optimal, such sets of items.

In this work, we present a more general, two-tiered approach for dealing with set preferences in the above setting. This approach combines a language for specifying certain types of set properties, and an *arbitrary* preference specification language for expressing preferences over single, attributed items. The basic idea is to first specify the set properties we care about, and then specify preferences over the values of these properties. Such a specification induces a preference ordering over sets based on the values these sets provide to the properties of interest. We believe that the suggested approach is both intuitive and powerful. Although in this paper we focus on a particular set of properties for which we have devised a relatively efficient optimization algorithm, in its most general form, this two-tiered approach generalizes the approach of desJardins and Wagstaff (2005) because diversity and specificity are just two set properties. In principle, one can express both more general properties referring to multiple attributes, as well as more general conditional preferences over the values of these properties.

---

1. We realize that common rules of rationality may not apply to users with such preferences.

Essentially, our approach re-states the problem of specifying preferences over sets in terms used to specify preferences over single items. In our formulation, "items" stand for the possible sets, and attributes of such "items" are their (user-defined) set-property values. Thus, in principle, this approach allows us to re-use any formalism for specifying preferences over single items. In this paper we will consider two specific instantiations of such a formalism: qualitative preferences based on CP or TCP-nets (Boutilier, Brafman, Domshlak, Hoos, & Poole, 2004; Brafman, Domshlak, & Shimony, 2006a), and quantitative preferences represented as generalized additively independent (GAI) value functions (Bacchus & Grove, 1995; Fishburn, 1969). The algorithm we suggest for computing an optimal subset given qualitative preferences is based on a similar optimization algorithm for TCP-nets. But because the number of "items" in our case is very large, this algorithm is modified substantially to exploit the special structure of these "items". These modifications enable us to compute an optimal subset faster.

## 2. Specifying Set Preferences

The formalism we use for set-preference specification makes one fundamental assumption: the items from which sets of interest are built are described in terms of some attributes, and the values of these attributes are what distinguishes different items. We shall use $\mathcal{S}$ to denote the set of individual items, and $\mathcal{X}$ to denote the set of attributes describing these items. For example, imagine that the "items" in question are US senate members, and the attributes and their values are: Party affiliation (Republican, Democrat), Views (liberal, conservative, ultra conservative), and Experience (experienced, inexperienced).

### 2.1 From Properties of Items to Properties of Item Sets

Given the set $\mathcal{X}$ of item-describing attributes, first, we can already talk about more complex item properties, e.g., "senate members with liberal views", or "inexperienced, conservative senate members". More formally, let $\overline{\mathcal{X}}$ be the union of the attribute domains, that is,

$$\overline{\mathcal{X}} = \{X = x \mid X \in \mathcal{X}, x \in Dom(X)\},$$

and let $\mathcal{L}_{\overline{\mathcal{X}}}$ be the propositional language defined over $\overline{\mathcal{X}}$ with the usual logical operators. $\mathcal{L}_{\overline{\mathcal{X}}}$ provides us with a language for describing complex properties of individual items. Since items in $\mathcal{S}$ can be viewed as models of $\mathcal{L}_{\overline{\mathcal{X}}}$, we can write $o \models \varphi$ whenever $o \in \mathcal{S}$ and $o$ is an item that satisfies the property $\varphi \in \mathcal{L}_{\overline{\mathcal{X}}}$.

Given the language $\mathcal{L}_{\overline{\mathcal{X}}}$, we can now specify arbitrary properties of item *sets* based on the attribute values of items in a set, such as the property of having at least two Democrats, or having more Democrats than Republicans. More generally, given any item property $\varphi \in \mathcal{L}_{\overline{\mathcal{X}}}$, we can talk about the number of items in a set that have property $\varphi$, which we denote by $|\varphi|(S)$, that is, $|\varphi|(S) = |\{o \in S | o \models \varphi\}|$. Often the set $S$ is implicitly defined, and we simply write $|\varphi|$. Thus, |Experience=experienced|$(S)$ is the number of experienced members in $S$. Often, we simply abbreviate this as |experienced|.

While $|\varphi|(\cdot)$ is an integer-valued property of sets, we can also specify boolean set properties as follows: $\langle |\varphi| \text{ REL } k \rangle$, where $\varphi \in \mathcal{L}_{\overline{\mathcal{X}}}$, REL is a relational operator over integers, and $k \in \mathbb{Z}^*$ is a non-negative integer. This property is satisfied by a set $S$ if $|\{o \in S | o \models \varphi\}|$ REL $k$. In our running example we use the following three set properties:

- $P_1 : \langle |\text{Party affiliation} = \text{Republican} \vee \text{Political view} = \text{conservative}| \geq 2 \rangle$

- $P_2 : \langle |\text{Experience} = \text{experienced}| \geq 2 \rangle$

- $P_3 : \langle |\text{Political view} = \text{liberal}| \geq 1 \rangle$

$P_1$ is satisfied (only) by sets with at least two members that are either Republican or conservative. $P_2$ is satisfied by sets with at least 2 experienced members. $P_3$ is satisfied by sets with at least one liberal.

We can also write $\langle |\varphi| \text{ REL } |\psi| \rangle$, with a similar interpretation. For example, $\langle |\text{Republican}| > |\text{Democrat}| \rangle$ holds for sets containing more Republicans than Democrats. An even more general language could include arithmetic operators (e.g., require twice as many Republicans as Democrats) and aggregate functions (e.g., the average number of years on the job). All these are instances of the general notion of specifying properties of sets as a function of the attribute values of the set's members. In this paper, we focus on the above language with the relational operators restricted to equalities and inequalities. We do so because having a clear, concrete setting eases the presentation, and because restricting the language allows us to provide more efficient subset-selection algorithms. Indeed, many of the ideas we present here apply to more general languages. In particular, this generality holds both for the overall preference-specification methodology, and for the search-over-CSPs technique for computing optimal subsets introduced later in the paper. However, the more specific techniques we use to implement these ideas, such as bounds generation, and the specific translation of properties into CSPs, rely heavily on the use of specific, more restrictive languages.

Finally, we note an important property of our preference specification approach of being independent of the actual set of items available at the moment. This generality is important for many applications where the same reasoning about set preferences must be performed on different, and often initially unknown sets of items. For example, this is the case with specifying guidelines for selecting articles for an online newspaper, or for selecting a set of $k$ results for an information query.

## 2.2 Reasoning with Set Preferences

Once we have specified the set properties of interest, we can define preferences over the values of these properties using any preference specification formalism. Here we discuss two specific formalisms, namely TCP-nets (Brafman et al., 2006a), an extension of CP-nets (Boutilier et al., 2004), and Generalized Additively Independent (GAI)-value functions (Bacchus & Grove, 1995; Fishburn, 1969). The former is a formalism for purely qualitative preference specification, yielding a partial preference order over the objects of interest. The latter is a quantitative specification formalism that can represent any value function.

Let $\mathcal{P} = \{P_1, \ldots, P_k\}$ be some collection of set properties. A TCP-net over $\mathcal{P}$ captures statements of the following two types:

(1) *Conditional Value Preference Statements.* "If $P_{i_1} = p_{i_1} \wedge \cdots \wedge P_{i_j} = p_{i_j}$ then $P_l = p_l$ is preferred to $P_l = p'_l$." That is, when $P_{i_1}, \ldots, P_{i_j}$ have a certain value, we prefer one value for $P_l$ to another value for $P_l$.

(2) *Relative Importance Statements.* "If $P_{i_1} = p_{i_1} \wedge \cdots \wedge P_{i_j} = p_{i_j}$ then $P_l$ is more important than $P_m$." That is, when $P_{i_1}, \ldots, P_{i_j}$ have a certain value, we prefer a better value for $P_l$ even if we have to compromise on the value of $P_m$.

Each such statement allows us to compare between certain pairs of item sets as follows:

- The statement **"if** $P_{i_1} = p_{i_1} \wedge \cdots \wedge P_{i_j} = p_{i_j}$ **then** $P_l = p_l$ **is preferred to** $P_l = p_l'$**"** implies that given any two sets $S, S'$ for which (1) $P_{i_1} = p_{i_1} \wedge \cdots \wedge P_{i_j} = p_{i_j}$ holds, (2) $S$ satisfies $P_l = p_l$ and $S'$ satisfies $P_l = p_l'$, and (3) $S$ and $S'$ have identical values on all properties except $P_l$, we have that $S$ is preferred to $S'$.

- The statement **"if** $P_{i_1} = p_{i_1} \wedge \cdots \wedge P_{i_j} = p_{i_j}$ **then** $P_l$ **is more important than** $P_m$**"** implies that given any two sets $S, S'$ for which (1) $P_{i_1} = p_{i_1} \wedge \cdots \wedge P_{i_j} = p_{i_j}$ holds, (2) $S$ has a more preferred value for $P_l$, and (3) $S$ and $S'$ have identical values on all attributes except $P_l$ and $P_m$, we have that $S$ is preferred to $S'$. (Notice that we do not care about the value of $P_m$ if $P_l$ is improved.)

We refer the reader to the work of Brafman et al. (2006a) for more details on TCP-nets, their graphical structure, their consistency, etc. The algorithms in this paper, when used with TCP-nets, assume an acyclic TCP-net Brafman et al.. The latter property ensures both consistency of the provided preferences, as well as existence of certain "good" orderings of $\mathcal{P}$ with respect to the TCP-net.

As an example, consider the following preferences of the president for forming a committee. He prefers at least two members that are either Republican or conservative, that is, he prefers $P_1$ to $\overline{P_1}$ unconditionally. (Depending on the context, we use $P$ to denote both the property $P$ and the value $P = true$. We use $\overline{P}$ to denote $P = false$.) If $P_1$ holds, he prefers $P_2$ over $\overline{P_2}$ (that is, at least two experienced members), so that the committee recommendations carry more weight. If $\overline{P_1}$ holds, he prefers $\overline{P_2}$ to $P_2$ (that is, all but one are inexperienced) so that it would be easier to influence their decision. The president unconditionally prefers to have at least one liberal, that is, he prefers $P_3$ to $\overline{P_3}$, so as to give the appearance of balance. However, $P_3$ is less important than both $P_1$ and $P_2$. There is an additional "external" constraint (or possibly a preference) that the total number of members be three.[2]

GAI value functions map the elements of interest (item sets in our case) into real values quantifying the relative desirability of these elements. Structure-wise, GAI value functions have the form $U(S) = \sum_{i=1,\ldots,n} U_i(\mathcal{P}_i(S))$, where each $\mathcal{P}_i \subset \mathcal{P}$ is a subset of properties. For example, the President's preferences imply the following GAI structure: $U(S) = U_1(P_1(S), P_2(S)) + U_2(P_3(S))$ because the President's conditional preferences over $P_2$'s value tie $P_1$ and $P_2$ together, but are independent of $P_3$'s value. $U_1$ would capture the weight of this conditional preference, combined with the absolute preference for $P_1$'s value. $U_2$ would represent the value of property $P_3$. We might quantify these preferences as follows: $U_1(P_1, P_2) = 10, U_1(P_1, \overline{P_2}) = 8, U_1(\overline{P_1}, P_2) = 2, U_1(\overline{P_1}, \overline{P_2}) = 5$; while $U_2(P_3) = 1, U_2(\overline{P_3}) = 0$. Of course, infinitely many other quantifications are possible.

---

2. Some external constraints, such as this cardinality constraint, can be modeled as a preference with high value/importance. In fact, this is how we model cardinality constraints in our implementation.

1: $\mathcal{Q} \leftarrow \{\emptyset\}$
2: $S_{opt} \leftarrow \emptyset$
3: **while** $\mathcal{Q}$ contains a set $S$ such that $\mathsf{UB}(S) > \mathsf{Value}(S_{opt})$ **do**
4: $\quad S \leftarrow \mathrm{argmax}_{S' \in \mathcal{Q}} \mathsf{UB}(S')$
5: $\quad \mathcal{Q} \leftarrow \mathcal{Q} \setminus \{S' \mid \mathsf{LB}(S_{opt}) \geq \mathsf{UB}(S')\}$
6: $\quad \mathcal{Q} \leftarrow \mathcal{Q} \bigcup \{S \cup \{o\} \mid o \in \mathcal{S} \setminus S\}$
7: $\quad S \leftarrow \mathrm{argmax}_{S' \in \mathcal{Q}} \mathsf{Value}(S')$
8: $\quad$ **if** $\mathsf{Value}(S) > \mathsf{Value}(S_{opt})$ **then**
9: $\quad\quad S_{opt} \leftarrow S$
10: $\quad$ **end if**
11: **end while**
12: **return** $S_{opt}$

Figure 1: Subset-space branch-and-bound search for an optimal subset of available items $\mathcal{S}$.

## 3. Finding an Optimal Subset

In general, given a preference specification and a set $\mathcal{S}$ of available items, our goal is to find an optimal subset $S_{opt} \subseteq \mathcal{S}$ with respect to the preference specification. That is, for any other set $S' \subseteq \mathcal{S}$, we have that the properties $S_{opt}$ satisfies are no less desirable than the properties $S'$ satisfies. We now consider two classes of algorithms for finding such an optimal subset. These two classes of algorithm differ in the space in which they search. In the next section, we describe a comparative empirical evaluation of these algorithms. For our running example we use the following set of available items $\mathcal{S}$:

| $o_1$ | Republican | conservative | inexperienced |
|-------|------------|--------------|---------------|
| $o_2$ | Republican | ultra conservative | experienced |
| $o_3$ | Democrat | conservative | experienced |
| $o_4$ | Democrat | liberal | experienced |

### 3.1 Searching in Sets Space

The most obvious approach for generating an optimal subset is to search directly in the space of subsets. A priori this approach is not too attractive, and indeed, we shall see later that our implementation of this approach did not scale up. However, given that often we are interested in sets of small size and that heuristics can be used to enhance search quality, we thought it is worth exploring this approach.

A branch-and-bound (B&B) algorithm in the space of sets is depicted in Figure 1. For each set $S$, the algorithm assumes access to an upper bound $\mathsf{UB}(S)$ and to a lower bound $\mathsf{LB}(S)$ estimates on the maximal value of a superset of $S$. The algorithm maintains a queue $\mathcal{Q}$ of sets, and this queue is initialized to contain only the empty set. At each step, the algorithm selects a highest upper-bound set $S$ from the queue. Next, the algorithm removes from $\mathcal{Q}$ all sets $S'$ with upper bound $\mathsf{UB}(S')$ being at most as good as the lower bound $\mathsf{LB}(S)$ of the selected set $S$, and adds to $\mathcal{Q}$ all the minimal (that is, one-item) extensions of $S$. The latter sets correspond to the successors of $S$ in the search space. Different implementations of the algorithm differ in how they sort the queue. The best-first version depicted in the pseudo-code sorts the queue according to a heuristic value of the set, and in our case this heuristic is an upper bound on the value of the set's supersets. In contrast, the depth-first version always positions the children of the newly expanded node at the front of the queue. We implemented and tested both versions.

The method used to generate bounds for a set $S$ must depend on the actual preference representation formalism, as well as on the type of set properties being used, and the idea is more natural given a quantitative value function. For a lower bound $\mathsf{LB}(S)$ we can use the actual value $\mathsf{Value}(S)$ of $S$. Note that it is possible that all descendants of $S$ will have lower values because, in general, set-properties may not be monotonic (e.g., "average value higher than 5.") However, since $S$ itself is a possible solution, this is a valid lower bound.

For an upper bound, we proceed as follows: First, we consider which set-property values are consistent with $S$. That is, for each set property, we examine what values $S$ and any of its supersets can potentially provide to that property. For example, consider $P_2$ and suppose $S$ contains a single experienced member. So currently, $\overline{P_2}$ holds. However, we can satisfy $P_2$ if we add one more experienced member. Thus, both values of $P_2$ are consistent with $S$. In contrast, if we had two experienced members in $S$, then $\overline{P_2}$ is inconsistent with $S$ because no matter who we add to $S$, we can never satisfy $\overline{P_2}$. Next, given such sets of possible set-properties' values with respect to the set $S$, we can bound the value of $S$ and of any of its supersets by maximizing values locally. Specifically, in a GAI value function, we can look at each local function $U_i$, and consider which assignment to it, from among the consistent values, would maximize $U_i$. Clearly, this may result in an overall value overestimation, since we do not know whether these "locally optimizing" joint assignments are consistent. Similar ideas can be used with other quantitative representations, as in various soft-constraint formalisms (Bistarelli, Fargier, Montanari, Rossi, Schiex, & Verfaillie, 1999).

Consider our running example with the GAI value function as at the end of Section 2, and consider searching for an optimal subset of $\mathcal{S} = \{o_1, o_2, o_3, o_4\}$ using a depth-first version of B&B. We start with the empty set, and the property values provided by the empty set are $\overline{P_1}, \overline{P_2}, \overline{P_3}$. Thus, the lower bound $\mathsf{LB}(\emptyset)$, which is the value of the empty-set, is 5. For the upper bound $\mathsf{UB}(\emptyset)$, we consider the best property values that are individually consistent with the extensions of $\emptyset$, which are $P_1, P_2, P_3$, and their accumulative value is 11. $S_{opt}$ is also initialized to the empty set, and next we generate all of the children of the (only possible) selected set $\emptyset$, which are all singleton sets: $\{o_1\}, \{o_2\}, \{o_3\}, \{o_4\}$. Except for $\{o_4\}$, they all have lower and upper bounds identical to those of the empty set, and are inserted into the queue. $\{o_4\}$ has a lower bound of 6 and the upper bound is 11. Suppose $\{o_1\}$ is the first queue element, and we select it for expansion. This results in adding $\{o_1, o_2\}, \{o_1, o_3\}, \{o_1, o_4\}$ into the queue, and the lower and upper bounds of these sets are $(8, 11), (8, 11), (6, 11)$, respectively. Next, the set $\{o_1, o_2\}$ is examined with respect to the current $S_{opt} = \emptyset$, and $S_{opt}$ is assigned to $\{o_1, o_2\}$. Since we assumed here a depth-first version of B&B we proceed with expanding $\{o_1, o_2\}$, obtaining $\{o_1, o_2, o_3\}, \{o_1, o_2, o_4\}$ with lower and upper bounds being, respectively, $(10, 11)$ and $(11, 11)$. With a lower bound of 11 for $\{o_1, o_2, o_4\}$ we can prune away all the rest of the nodes in the queue, and we are done.

An important issue for depth-first B&B is the order in which sets are generated. In our implementation, at each node in the search space, the items in $\mathcal{S}$ are ordered according to the sum of the value of the properties they can help satisfy. For example, initially, a conservative member such as $o_1$ could help us satisfy $P_1$.

In contrast to quantitative preference representation formalisms, qualitative preferences typically induce a partial ordering over property collections. In this case, it is harder to generate strict upper and lower bounds – as they must be comparable to any possible solution. One way to handle this is to linearize the ordering and require the stronger property of optimality with respect to the resulting total order. Here, TCP-nets present themselves as a good choice because there is an efficient and simple way of generating a value function consistent with an acyclic TCP-net (Brafman & Domshlak, 2008). This value function retains the structure of the original network which is important to make the bounds computation efficient (notably, each $U_i$ depends on a small number of property values).

## 3.2 Searching over CSPs

The attractiveness of the item subsets is evaluated in terms of a fixed collection of set-properties $\mathcal{P}$, and thus different sets that provide all identical property values are equivalent from our perspective. The immediate conclusion is that considering separately such preferentially equivalent subsets of available items $\mathcal{S}$ is redundant. To remove this redundancy, we suggest an alternative method in which we search directly over set-property value combinations. Of course, the problem is that given a set-property value combination, it is not obvious whether we can find an actual subset of $\mathcal{S}$ that has such a combination of properties. To answer this question, we generate a CSP that is satisfiable if and only if there exists a subset of $\mathcal{S}$ with the considered set-property values. The overall search procedure schematically works as follows.

1. Systematically generate combinations of set-property values.

2. For each such combination, search for a subset of $\mathcal{S}$ providing that combination of set-property values.

3. Output a subset of $\mathcal{S}$ satisfying an optimal (achievable) combination of set-property values.

To make this approach as efficient as possible, we have to do two things, namely:

(1) Find a way to prune sub-optimal set-property value combinations as early as possible.

(2) Given a set-property value combination, quickly determine whether a subset of $\mathcal{S}$ satisfies this combination.

Considering the first task, let $P_1, \ldots, P_k$ be an ordering of the set-properties $\mathcal{P}$.[3] Given such an ordering of $\mathcal{P}$, we incrementally generate a tree of property combinations. The root of that tree corresponds to an empty assignment to $\mathcal{P}$. For each node $n$ corresponding to a partial assignment $P_1 = p_1, \ldots, P_j = p_j$, and for every possible value $p_{j+1}$ of the property $P_{j+1}$, the tree contains a child of $n$ corresponding to the partial assignment $P_1 = p_1, \ldots, P_j = p_j, P_{j+1} = p_{j+1}$. The tree leaves correspond to (all) complete assignments to $\mathcal{P}$. Such a tree for our running example is depicted in Figure 2. Note that, implicitly, each node in this tree is associated with a (possibly empty) set of subsets of $\mathcal{S}$, notably, the subsets that provide the set-property value combination associated with that node.

In our search for an optimal set, we expand this tree of set-property value combinations while trying to expand as few tree nodes as possible by pruning certain value combinations of $\mathcal{P}$ as either sub-optimal with respect to set preferences, or unsatisfiable with respect to $\mathcal{S}$. A standard way to do this is, again, by using a branch-and-bound search procedure, and this requires from us to derive effective upper and lower bounds on the value of the best subset satisfying a partial value combination for $\mathcal{P}$. In addition, the order we associate with properties and their values affects our pruning ability throughout the search process. To get the most leverage out of our bounds, we would like to explore the children of a node in the decreasing order of their purported attractiveness. Moreover, when fixing the ordering of the set-properties themselves, we would like properties that can potentially contribute more to appear earlier in this ordering. For instance, $P_1$'s value in our running example has a greater influence on the overall attractiveness of a subset than the value of $P_2$, and thus $P_1$ should better be branched on first. In addition, $P_1$ is preferred to be *true*, and thus the subtree corresponding to $P_1 = true$ should better be explored first. Similarly, $P_2$ is preferred to be *true* when $P_1 = true$, and preferred to be *false*, otherwise. This ordering is reflected in the tree in Figure 2, for a left to right pre-order traversal of the tree.

---

3. Throughout this paper, we will assume that in preference specifications using TCP nets, there are only conditional preference (CP) arcs, and importance arcs, but no conditional importance (CI) arcs. While our scheme and implementations allow these arcs, CI arcs force the ordering of the set properties to be dynamic, as it may depend on value assignments to previous properties. For clarity of exposition, we thus preferred not to present these technical details.

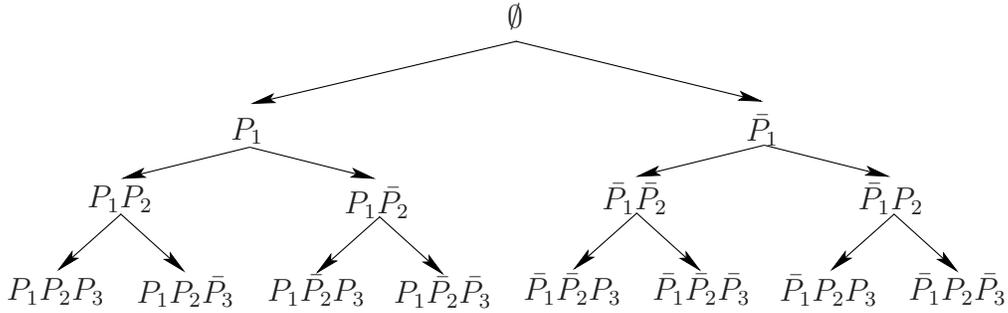

Figure 2: Illustration of a search tree for our running example.

Now, let us consider the second task of determining whether a subset of $\mathcal{S}$ satisfies a given set-property value combination. Given such a partial assignment $\alpha$ to $\mathcal{P}$, we set up the following CSP. First, the CSP has a boolean variable $x_i$ for every available item $o_i \in \mathcal{S}$. In our example, the CSP contains the variables $x_1, \dots, x_4$ for items $o_1, \dots, o_4$ respectively. Intuitively, $x_i = 1$ encodes $o_i$ being a part of our (searched for) subset of $\mathcal{S}$, whereas $x_i = 0$ means that $o_i$ is not in that subset. Next, we translate every set-property value in $\alpha$ into a certain constraint on these variables. For instance, if $\alpha[P_1] = true$, the constraint $C_1 : x_1 + x_2 + x_3 \geq 2$ is added to the CSP. Note that $C_1$ explicitly encodes the requirement (of $P_1 = true$) for the subset to have at least two of the elements that satisfy Republican $\vee$ conservative. That is because $\{o_1, o_2, o_3\}$ are all the candidates in $\mathcal{S}$ that are either Republican or conservative. Alternately, if $\alpha[P_1] = false$, then the constraint $\overline{C_1} : x_1 + x_2 + x_3 < 2$ is added to the CSP. Finally, if $\alpha$ does not specify a value for $P_1$, then no constraints related to $P_1$ should be added at all. Likewise, for $\alpha[P_2] = true$ and $\alpha[P_3] = true$ we would add the constraints $C_2 : x_2 + x_3 + x_4 \geq 2$ and $C_3 : x_4 \geq 1$, respectively. In general, it is not hard to verify that the CSP constructed this way for a concrete item set $\mathcal{S}$ and a set-property value combination $\alpha$ is solvable if and only if $\mathcal{S}$ has a subset satisfying $\alpha$. Moreover, if this CSP is solvable, then any of its solutions explicitly provides us with such a subset of $\mathcal{S}$.

It is worth briefly pointing out the difference between the CSPs we generate here and the more typical CSPs usually discussed in the literature. Most work on general CSPs deals with constraints over small, typically just two-variable, subsets of problem variables. In contrast, the constraints in the CSPs generated in our optimization process are global, with each constraint being possibly defined over *all* the CSP variables. Yet another special property of CSPs constructed for our purposes is that there is a sense in which it is meaningful to talk about partial assignments in our context—unassigned variables can always be regarded *de facto* as if assigned the value "0" since the corresponding items, by default, do not belong to the subset we search for.

Because partial assignments to set-properties $\mathcal{P}$ map to CSPs, each node in our tree of set-property value combinations maps to a CSP, and the entire tree can be viewed as a tree of CSPs. The important property of this tree-of-CSPs is that the children of each CSP node are CSPs obtained by adding one additional constraint to the parent CSP, notably the constraint corresponding to the additional property value that we want the set to satisfy. This implies that if some CSP node in the tree is unsatisfiable, then all of its descendants are unsatisfiable as well. In fact, we can make a stronger use of the nature of this search tree, recognizing that we can reuse the work done on a parent node to speed up the solution of its children. To see the latter, consider some CSP $\mathscr{C}$ in our tree-of-CSPs, some child CSP $\mathscr{C}'$ of $\mathscr{C}$, and let $S \subseteq \mathcal{S}$ be a solution to $\mathscr{C}$. As $\mathscr{C}'$ extends $\mathscr{C}$ with a constraint $C$, any subset $S' \subseteq \mathcal{S}$ ruled out by $\mathscr{C}$ will be also ruled out by $\mathscr{C}'$. Hence, if solving $\mathscr{C}$ and $\mathscr{C}'$ considers subsets of $\mathcal{S}$ in the same order (that is, by using the same ordering over set elements), then solving $\mathscr{C}'$ can start from the leaf node corresponding to $S$, the solution generated for $\mathscr{C}$. Moreover, if a constraint $C$ represents a boolean set property, and $S$ is not a solution to $\mathscr{C}' = \mathscr{C} \cup \{C\}$, then $S$ has to be a solution to $\mathscr{C} \cup \{\neg C\}$, which is the sibling of $\mathscr{C}'$. Using these ideas, we share the work done on different CSP nodes of our tree-of-CSPs. In fact, when all set properties are boolean, this approach needs to backtrack over each property at most once (we call this property "limited backtracking"), thereby considerably improving the empirical performance of the algorithm.

The overall branch-and-bound algorithm in the space of CSPs is depicted in Figure 3. As is, the algorithm is formulated for the case of quantitative preference formalisms. The formulation of the algorithm for the qualitative case is essentially the same, with minor technical differences and an important computational property. For CP/TCP-nets, we can guarantee that only limited backtracking is required if we follow the following guidelines. First, we must order the variables (line 1) in an order consistent with the topology of the network. Note that for TCP-nets, this ordering may be conditional, that is, the order of two variables may vary depending on the value of some of the earlier variables. Second, in line 2, the property values must be (possibly partially) ordered from best to worst, given the values of the parent properties (which must be and will be instantiated earlier). In that case, the first satisfiable set of properties constitutes an optimal choice (Brafman et al., 2006a). Assuming we solve intermediate nodes in the tree-of-CSPs, we know that we should backtrack at most once in each level assuming boolean set-properties, but, again, more backtracks may occur with integer-valued properties.

The node data structure used by the algorithm has two attributes. For a search node $n$,

- $n.\alpha$ captures a partial assignment to the set-properties $\mathcal{P}$ associated with the node $n$, and

- $n.S$ captures a subset of $\mathcal{S}$ satisfying $n.\alpha$ if such exists, and otherwise has the value *false*.

The functions Value, LB, and UB have the same semantics as in the subset-space search algorithm in Figure 1. In the pseudocode we assume a fixed ordering over set-property values (line 2), but one can vary it depending on earlier values (and we exploit that in our implementation). Finally, the

1: Fix an ordering over set-properties $\mathcal{P}$
2: Fix an ordering over the values of each set property $P \in \mathcal{P}$
3: Fix an ordering over all available items $\mathcal{S}$
4: $\mathcal{Q} \leftarrow \{n[\emptyset; \emptyset]\}$
5: $S_{opt} \leftarrow \emptyset$
6: **while** $\mathcal{Q}$ is not empty **do**
7:     $n \leftarrow pop(\mathcal{Q})$
8:     construct-and-solve-csp($n$)
9:     **if** $n.S \neq false$ and UB($n.S$) > Value($S_{opt}$) **then**
10:         **if** Value($n.S$) > Value($S_{opt}$) **then**
11:             $S_{opt} \leftarrow n.S$
12:         **end if**
13:         Let $P$ be the highest-ordered set property unassigned by $n.\alpha$
14:         **for each** possible value $p$ of $P$ **do**
15:             $n' \leftarrow [n.\alpha \cup \{P = p\}; n.S]$
16:             $\mathcal{Q} \leftarrow \mathcal{Q} \cup \{n'\}$             $\triangleright$ The position of $n'$ in $\mathcal{Q}$ depends on the search strategy
17:         **end for**
18:     **end if**
19: **end while**
20: **return** $S_{opt}$

Figure 3: CSP-space branch-and-bound search for an optimal subset of available items $\mathcal{S}$.

pseudo-code leaves open the choice of search strategy for used by the branch-and-bound, and this choice is fully captured by the queue insertion strategy in line 16.

To illustrate the flow of the algorithm, let us consider again our running example. Recall that the example already has a requirement for the discovered subset to be of size 3, and this translates into a constraint $C : x_1 + x_2 + x_3 + x_4 = 3$. The first CSP we consider has $\{C, C_1\}$ as its only constraints. Assume the CSP variables are ordered as $\{x_1, x_2, x_3, x_4\}$, with value 1 preceding value 0 for all $x_i$. In that case, the first solution we find is $S_1 : x_1 = 1, x_2 = 1, x_3 = 1, x_4 = 0$. Our next CSP adds the constraint $C_2$. When solving this CSP, we continue to search (using the same order on the $x_i$'s and their values) from the *current* solution $S_1$, which turns out to satisfy $C_2$ as well. Thus, virtually no effort is required to solve this CSP. Next, we want to also satisfy $C_3$. This set of constraints corresponds to a leaf node in the tree-of-CSPs which corresponds to the complete assignment $P_1 P_2 P_3$ to the set-properties. Our current item set $S_{opt} = S_1$ does not have a liberal, so we have to continue to the assignment $S_2 : x_1 = 1, x_2 = 1, x_3 = 0, x_4 = 1$ (requiring us to backtrack in the CSP-solution space over the assignments to $x_4$ and $x_3$). We now have a set that satisfies the properties in the leftmost leaf node in our tree-of-CSPs. If we can prove that this set-property value combination is optimal using our upper/lower bounds, we are done. Otherwise, we need to explore additional nodes in our tree-of-CSPs. In the latter case, the next CSP will correspond to $P_1, P_2, \overline{P_3}$, with constraints $\{C, C_1, C_2, \overline{C_3}\}$. However, we already have a solution to this node, and it is exactly $S_1$. To see that, note that $S_1$ was a solution to the parent of our current CSP, but it was not a solution to its sibling $\{C, C_1, C_2, C_3\}$. Hence, since $P_3$ is a boolean property, $S_1$ must satisfy $\{C, C_1, C_2, \overline{C_3}\}$.

### 3.3 Solving the underlying CSPs

Our algorithm for solving the intermediate CSPs is based on the well known backtrack-search algorithm, first presented by Prosser (1993) in a simple iterative form. At the same time, we have adapted both the algorithm and some well known enhancements in CSP solving (such as NoGood recording and forward checking (FC)) to the specifics of the CSPs in our setting.

Initially, variables and their values are statically ordered from the most to least constrained (although we also discuss a few experiments performed with dynamic variable/value ordering). Our motivation for static ordering is two-fold. First, because the constraints are very much global, we can do the ordering at a preprocessing stage. Second, as discussed in the previous section, static ordering allows us to better utilize solutions of CSPs when solving descendent CSPs.

The basic backtrack algorithm, which on its own, unsurpisingly performs quite poorly in our setting, is refined by utilizing the following observations and techniques.

- **Monotonicity of improving constraints.** If the operator of the constraint is "=" and there are more items having the constrained property already in the current partial solution, then one cannot satisfy the constraint by making additional assignments. The same property holds for the constraint operators "<" and "≤". Using this observation, it is possible to detect the need to backtrack early on in the search.

- **Forward Checking.** A certain type of "forward checking" can be performed for our constraints. Clearly, if satisfying some constraint requires at least $k$ items to be added to the subset, and the number of remaining items that satisfy the desired property is less than $k$, then the search algorithm must backtrack.

- **"Can/Must" strategy.** The "can/must" strategy corresponds to a more advanced check of the interactions between the constraints. The idea is quite simple: if (i) *at least* $p$ items *must* be added to the constructed subset to satisfy the constraint $C_i$, (ii) *at most* $q$ items *can* be added to the constructed subset without violating another constraint $C_j$, (iii) all the items that can be added and have the property constrained by $C_i$ also have the property constrained by $C_j$, and, finally, (iv) $p > q$, then both $C_i$ and $C_j$ cannot be satisfied simultaneously. Moreover, no further assignments to yet unassigned variables can resolve this conflict, and thus the situation is a dead end. This kind of reasoning allows discovery of such barren nodes quite early in the search, pruning large portions of the search tree. To reason correctly about the "can/must" strategy, we have to maintain a data structure of unique items for each pair of constraints, as well as to keep track of the number of remaining items that influence property constrained by $C_i$ and do *not* influence properties constrained by $C_j$.

  As an example, assume we are in the middle of the search and we have two set properties: $SP_1 : |A_1 = a| \geq 5$ and $SP_2 : |A_2 = b| \leq 3$. Suppose that we have already picked 3 items that influence $SP_1$ and 2 items that influence $SP_2$. As a result, to satisfy $SP_1$, we *must* add at least another two items that influence it and to satisfy $SP_2$ we *can* add at most one item that influences $SP_2$. If all the items that we can choose from $\{o_k...o_n\}$ have a value "a" for the attribute $A_1$ and value "b" for the attribute $A_2$, then obviously we cannot satisfy both $SP_1$ and $SP_2$ within this setting, and thus we should backtrack.

Finally, below we discuss recording NoGoods, an improvement of the basic backtracking algorithm that proved to have the most impact in our setting.

### 3.3.1 NOGOOD RECORDING

The standard definition of a NoGood in the CSP literature is that of a partial assignment that cannot be extended into a full solution of the problem. Once we learn a NoGood, we can use it to prune certain paths in the search tree. The smaller the NoGood, the more occasions we can use it, the greater its pruning power. Thus, it is of interest to recognize minimal NoGoods, and different techniques have been developed to perform NoGood resolution in order to produce the best and most general NoGoods possible (see, e.g., Dechter, 1990; Schiex & Verfaillie, 1993; Dago & Verfaillie, 1996).

As noted earlier, the CSPs we generate differ significantly from the more typical binary CSPs. Consequently, the NoGood recording algorithm has to be adapted accordingly. In particular, because our constraints are global, it makes sense to try generating NoGoods that are global, too. Thus, instead of recording assignments to variables, we record the influence of the current assignment on the constraints. Every variable influences a set of constraints.[4] Thus, as a NoGood, we store the influence the set selected so far has on all the constraints. Specifically, suppose we have generated the set $S_1$, and recognized that it is not extensible into a set satisfying the constraints. (This immediately follows from the fact that we backtracked over this set.) We now generate a No-Good $N$ that records for each property associated with each constraint, how many items satisfying that property occur in $S_1$. Now, suppose we encounter a different set $S_2$ that has the same effect $N$ on the constraints. If there are fewer options to extend $S_2$ than there are to extend $S_1$, we know that $S_2$, as well, cannot be extended into a solution. However, if there are more options to extend $S_2$ than $S_1$, we cannot conclude that $S_2$ is a NoGood at this point. In order to better quantify the options that were available to extend $S_1$ we record, beyond the actual NoGood $N$, the level (depth) in the assignment tree at which it was generated. Given that the CSP solver uses a static variable ordering, we know that if we encounter a set $S$ that generates the same properties as the NoGood $N$, at a level no higher than that of $S_1$, we can safely prune its extensions. The reason for that is, there are no additional extension options available for $S$ than there were for $S_1$.

The correctness of the NoGood recording mechanism proposed here depends on having a static variable ordering, as well as a specific value ordering for all the variables in the CSP, namely, $\langle 1, 0 \rangle$. To show correctness, we should note that a NoGood can be used only after it is recorded. Consequently, any node using a NoGood would be to the right in the search tree of a node the NoGood was recorded at. Here we would like to stress again that, since the constraints are global, it does not matter which items are added to the subset, but rather what influence these items had on the constraints. Any two sets having exactly the same influence on the constraints are identical with respect to the optimization process.

### 3.3.2 SEARCH ALGORITHM

The procedure depicted in Figure 4 extends the basic backtrack algorithm by a subroutine CANIM-PROVE which can be altered to include any combination of the in-depth checks discussed earlier, to utilize early conflict detection techniques, including the NoGoods check. Also added is a call to the ADDNOGOOD subroutine for recording NoGoods while backtracking. $P$ and $n$, the generated instance of a CSP problem with variables indexed from 1 to $|\mathcal{S}|$ and the node in the tree-space search respectively, are the inputs to the procedure. The algorithm systematically tries to assign values to the problem variables, backtracking and recording NoGoods when facing a dead end.

---

4. We assume without loss of generality that every item in the set of available items influences at least one constraint in the constraint set $\mathscr{C}$, since items that influence no constraint can be safely eliminated.

```
1:  consistent ← n.S satisfies n.α
2:  while not(consistent) do
3:      if HASVALUES(P.vars[i]) and  CANIMPROVE(P)  then
4:          P.i ← LABEL(P.i, consistent)    ▷ If current CSP variable has available values, try to set, update consistency
5:      else
6:           ADDNOGOOD(P,i)                                                          ▷ Record NoGood
7:          P.i ← UNLABEL(P.i)                                                       ▷ Backtrack
8:      end if
9:      if P.i = 0 then                     ▷ If backtracked over the first indexed variable — no solution available
10:         return false
11:     end if
12: end while
13: return true
```

Figure 4: Conflict backtrack algorithm with NoGood recording

## 4. Experimental Results

We evaluate the different algorithms using a subset of the movie database publicly available from `imdb.com`. We simulated a scenario of selecting movies for a three-day film festival according to organizers preferences. Three models of growing complexity have been engineered to reflect the preferences of the organizers; these models are defined in terms of 5, 9, and 14 set-properties, respectively. In addition, the total number of films is constrained to be 5 (which we actually modeled using a very strong preference). Figure 5 depicts the list $\mathcal{P}_{14}$ of the 14 properties and their alterations; $\mathcal{P}_5$ and $\mathcal{P}_9$ consist of the corresponding prefixes ($SP_1$ through $SP_5$, and $SP_1$ through $SP_9$, respectively) of $\mathcal{P}_{14}$. To produce even more complex problem instances that cause many backtracks in the space of set-property assignments we slightly altered the 14-properties model, creating two additional models that are denoted henceforth as $\mathcal{P}'_{14}$ and $\mathcal{P}''_{14}$.

### 4.1 Preference Specification

Figure 6 provides a verbal description of qualitative preferences for the film festival program which we used in our experiments. Figure 7 depicts a TCP-net that encodes these preferences in terms of the more concrete set-properties listed in Figure 5. For the experiments with GAI value functions, these preferences were quantified by compiling this TCP-net into a GAI value function that orders the items consistently with that TCP-net (Brafman & Domshlak, 2008). The task in our empirical evaluation was to find an optimal subset of a set of available movies $\mathcal{S} \in \{\mathcal{S}_{400}, \mathcal{S}_{1000}, \mathcal{S}_{1600}, \mathcal{S}_{3089}\}$, where $\mathcal{S}_i$ corresponds to a set of $i$ movies, and that with respect to each of the five models of preferences over sets. All the experiments were conducted using Pentium 3.4 GHz processor with 2GB memory running Java 1.5 under Windows XP Professional. The runtimes reported in the tables below are all in seconds, with "–" indicating process incompletion after four hours.

$SP_1 = \langle |\text{Year} \geq 2002| = 5 \rangle$

$SP_2 = \langle |\text{Genre} = \text{Comedy}| \geq 2 \rangle$

$SP_3 = \langle |\text{Genre} = \text{Thriller}| \leq 3 \rangle$

$SP_4 = \langle |\text{Genre} = \text{Family}| > 1 \rangle$

$SP_5 = \langle |\text{Color} = \text{B\&W}| > 1 \rangle$

$SP_6 = \langle |\text{Director} = \text{Spielberg}| \geq 1 \rangle$

$SP_6{}^* = \langle |\text{Director} = \text{Spielberg}| \leq 1 \rangle$

$SP_7 = \langle |\text{Sound} = \text{Mono}| \geq 2 \rangle$

$SP_8 = \langle |\text{Genre} = \text{War} \ \vee \ \text{Genre} = \text{Film-noir}| = 0 \rangle$

$SP_8{}^* = \langle |\text{Genre} = \text{War} \ \vee \ \text{Genre} = \text{Film-noir}| \geq 4 \rangle$

$SP_8{}^{**} = \langle |\text{Genre} = \text{Film-noir}| \geq 4 \rangle$

$SP_9 = \langle |\text{Location} = \text{North America}| > 1 \rangle$

$SP_{10} = \langle |\text{Actor} = \text{Famous} \ \vee \ \text{Actress} = \text{Famous}| = 5 \rangle$

$SP_{11} = \langle |\text{Actress} = \text{Famous}| \geq 2 \rangle$

$SP_{12} = \langle |\text{Genre} = \text{Drama}| \geq 2 \rangle$

$SP_{13} = \langle |\text{Release Date} < 1970| \leq 1 \rangle$

$SP_{14} = \langle |\text{Net Profit} \geq 1000000| \geq 2 \rangle$

$SP_{14}{}^{**} = \langle |\text{Net Profit} \geq 1000000| \geq 5 \rangle$

Figure 5: Set-properties used in modeling user preferences in the movies selection domain.

$^*$ Alteration of $\mathcal{P}_{14}$, to achieve more backtracking - denoted as $\mathcal{P}'_{14}$

$^{**}$ Further alteration of $\mathcal{P}'_{14}$ to achieve even more backtracking - denoted as $\mathcal{P}''_{14}$

1. I prefer new movies to old movies, and therefore prefer that all movies be from 2002 or later, and this is important to me.

2. I love comedies, thrillers and family movies.

3. I prefer not to have too many movies in black and white (not more than one such movie).

4. If all the movies are new (after 2002) then I would prefer to have at least 2 comedies.

5. If I can find at least 2 comedies then I also prefer to have more than 1 family movie, but less then 3 thrillers. However having the right number of family movies is more important to me than having the right number of thrillers.

6. If not all the movies are new, I prefer to have at least 2 movies in black and white for the vintage touch.

7. If not all the movies are new, I prefer at least one movie to be directed by Steven Spielberg, but otherwise, I don't like his newer films

8. If the previous condition holds, then the number of movies with mono sound may be greater than 2.

9. I prefer not to have any war films or film-noir in the festival. However if this condition can not be satisfied, then I prefer not to have any films that were filmed in North America and this is more important to me than my preferences about the movie being in color or in B&W.

10. To draw more attention, I prefer all 5 movies to have famous actors or actresses.

11. To highlight female roles, I prefer at least 2 movies with a famous actress.

12. I prefer to have at least 2 dramas because people tend to think dramas are more sophisticated movies than any other genre.

13. I prefer to have at least one classical movie.

14. I prefer to have at least one commercially successful movie, i.e. a movie whose net profit was more than one million dollars.

Figure 6: Informal description of the assumed preferences for selecting a set of movies for a film festival program.

First, our initial experiments quickly showed that the search in the space of subsets (Table 1) does not scale up. With just over 20 elements, it did not converge to an optimal solution within an hour, even when the preference specification involved only 5 set-properties. This outcome holds for all combinations of qualitative and quantitative preference specifications, depth-first and best-first schemes of branch-and-bound, and queue ordering based on set's upper bound, lower bound, and weighted combinations of both. Table 1 provides a snapshot of the corresponding results for a TCP-net specified over nine set properties. The table describes the total number of subsets generated until an optimal subset was found (see the column "Subset until $S_{opt}$"), the total number of subsets generated until the optimal subset was recognized as optimal (under "Subsets generated"). DFS appears to be much more effective than BFS, but the branching factor of larger databases overwhelms this approach. Also, it may be thought that with larger databases it should be easier to quickly generate good sets, but we found that for moderately larger (e.g.,. 25+) and much larger (e.g., 3000) datasets, this approach is too slow. Various improvements may be possible, but given the much better performance of the other approach discussed later, they are unlikely to make a difference.

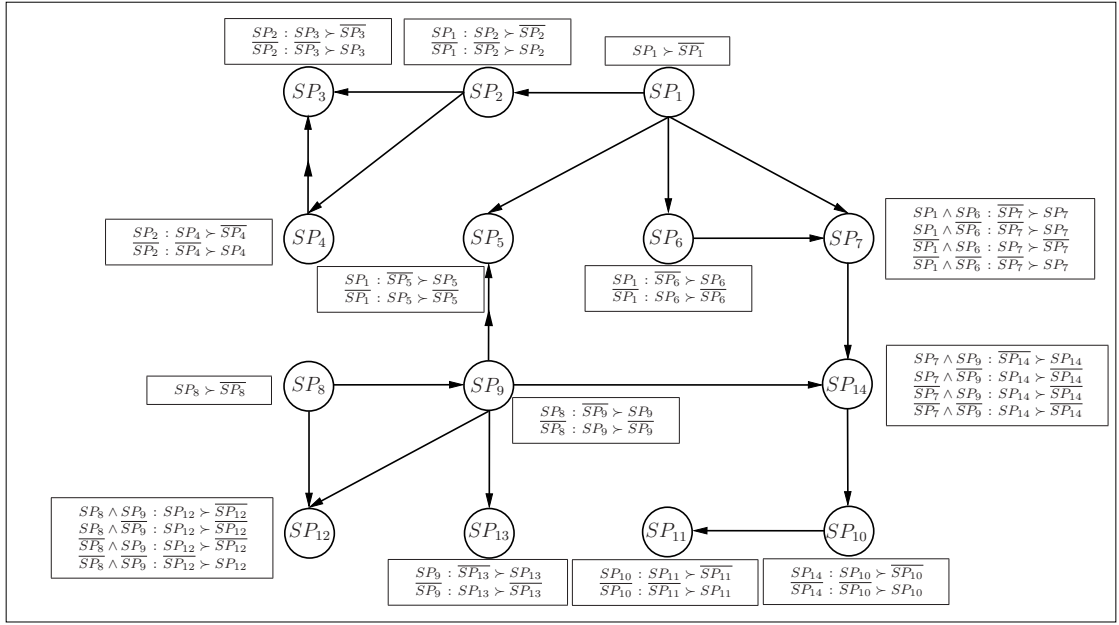

Figure 7: TCP-net model of preference over sets of movies for the film festival program.

Next, we consider the CSP-space branch-and-bound search. In particular, here we compared between the two variants of this approach that use dynamic and static variable and value orderings. In what follows, these two variants are denoted as **BB-D** and **BB-S**, respectively. While static variable/value orderings are usually considered to be a weaker approach to CSP solving, earlier we have shown that, in our domain, static ordering allows for certain optimizations that have a potential to improve the efficiency of the overall problem solving. In particular, static variable ordering allows to record global NoGoods as described in Section 3.3.1; the results for algorithms that record NoGoods are denoted by a name suffix "**+ng**". In addition, we have tried to share information between consecutive CSP problem instances while doing the search in the tree of CSPs; the algorithms adopting this technique are denoted by a name suffix "**+inc**".

| | Method | Subsets generated | Subsets until $S_{opt}$ | Time (sec) |
|---|---|---|---|---|
| $S_8$ | BFS | 4075 | 18 | 0.56 |
| $S_8$ | DFS | 630 | 83 | 0.19 |
| $S_{10}$ | BFS | 15048 | 40 | 2.34 |
| $S_{10}$ | DFS | 2935 | 672 | 0.47 |
| $S_{15}$ | BFS | 104504 | 7879 | 68.23 |
| $S_{15}$ | DFS | 30547 | 11434 | 3.13 |
| $S_{20}$ | BFS | 486079 | 28407 | 1584.67 |
| $S_{20}$ | DFS | 231616 | 28407 | 28.578 |

Table 1:  A snapshot of the results for subsets-space search. The preferences here are specified by a TCP-net over nine set properties.

| Set-properties | Method | $S_{400}$ | $S_{1000}$ | $S_{1600}$ | $S_{3089}$ |
|---|---|---|---|---|---|
| $P_5$ | **BB-D** | 0.3 | 0.77 | 1.30 | 4.02 |
| $P_5$ | **BB-S** | 0.14 | 0.14 | 0.17 | 0.25 |
| $P_5$ | **BB-S+inc** | **0.05** | **0.1** | **0.12** | **0.18** |
| $P_5$ | **BB-S+ng** | 0.17 | **0.1** | 0.15 | 0.21 |
| $P_5$ | **BB-S+ng+inc** | **0.05** | 0.11 | 0.13 | 0.19 |
| $P_9$ | **BB-D** | 0.43 | 1.42 | 2.42 | 6.58 |
| $P_9$ | **BB-S** | 0.14 | 0.24 | 0.26 | 0.34 |
| $P_9$ | **BB-S+inc** | **0.06** | **0.14** | **0.17** | **0.15** |
| $P_9$ | **BB-S+ng** | 0.17 | 0.25 | 0.34 | 0.35 |
| $P_9$ | **BB-S+ng+inc** | **0.06** | **0.14** | 0.18 | 0.17 |
| $P_{14}$ | **BB-D** | 0.66 | 2.03 | 4.69 | 14.92 |
| $P_{14}$ | **BB-S** | 0.17 | 0.43 | 1.09 | 0.78 |
| $P_{14}$ | **BB-S+inc** | **0.06** | **0.15** | 0.43 | **0.5** |
| $P_{14}$ | **BB-S+ng** | 0.3 | 0.57 | 1.06 | 0.95 |
| $P_{14}$ | **BB-S+ng+inc** | 0.1 | 0.19 | **0.38** | 0.54 |
| $P'_{14}$ | **BB-S** | 6.5 | 27.1 | 278 | – |
| $P'_{14}$ | **BB-S+inc** | **2.1** | 27 | 259 | – |
| $P'_{14}$ | **BB-S+ng** | 16.1 | 19.4 | **54.8** | 230.2 |
| $P'_{14}$ | **BB-S+ng+inc** | 4.68 | **18.4** | 76.3 | **210.8** |
| $P''_{14}$ | **BB-D** | 4113.48 | – | – | – |
| $P''_{14}$ | **BB-S** | 101.4 | 5370 | 16306 | – |
| $P''_{14}$ | **BB-S+inc** | **81.03** | 5523 | 16643 | – |
| $P''_{14}$ | **BB-S+ng** | 110 | 269.9 | **646.1** | 3335 |
| $P''_{14}$ | **BB-S+ng+inc** | 107.9 | **266.8** | 646.8 | **3013** |

Table 2:  Empirical results of evaluating the CSP-space search procedures with qualitative preference specification using TCP-nets.

Table 2 depicts the results of the evaluation of all variants of the CSP-space branch-and-bound search algorithm (Figure 3). First, the table shows that the overhead of maintaining NoGoods does not pay off for the simple preference specifications. However, for the more complex problems requiring more intense CSP solving, the use of NoGood recording proved to be very useful, letting us solve previously unsolvable instances. Next, the reader may notice from the table that, at least for the problems used in our tests, the contribution of the incremental approach is not substantial. For instance, NoGood recording by itself seems to contribute much more to the efficiency of the optimization process. Moreover, for the more complex problems, switching to the incremental version sometimes even leads to performance degradation. It appears that the overhead of maintaining and copying the partial solution in these cases does not pay off.

| Set-properties | Method | $\mathcal{S}_{400}$ | $\mathcal{S}_{1000}$ | $\mathcal{S}_{1600}$ | $\mathcal{S}_{3089}$ |
|---|---|---|---|---|---|
| $\mathcal{P}_5$ | **BB-S** | 0.24 | 0.14 | 0.17 | 0.26 |
| $\mathcal{P}_9$ | **BB-S** | 0.16 | 0.25 | 0.28 | 0.41 |
| $\mathcal{P}_{14}$ | **BB-S**+inc | 38.91 | 2376.40 | – | – |
| $\mathcal{P}_{14}$ | **BB-S**+ng+inc | 19 | 160.53 | 494.98 | 1349.9 |

Table 3: Results for the CSP-space search with quantitative preference specification using GAI value functions.

Our next set of experiments mirrored the first one, but now with GAI value functions instead of the purely qualitative TCP-nets. The GAI functions were obtained by properly quantifying the qualitative preferences used for the first tests. Table 3 provides a representative snapshots of the results. With value functions over set-properties $\mathcal{P}_5$ and $\mathcal{P}_9$ the basic branch-and-bound algorithm with static variable/value orderings performs and scales up (with growing set of alternatives $\mathcal{S}$) quite well. With the more complex value functions over the larger set of properties $\mathcal{P}_{14}$ the performance significantly degrades, and even the incrementality-enhanced algorithm cannot solve problem instances with more than 1000 CSP variables. On the other hand, adding NoGoods recording proves to dramatically improve the performance, leading to solving even the largest problem instances.

Tables 2 and 3 suggest a qualitative difference in the performance of the CSP-space search with quantitative and qualitative preference representation models. There are good reasons to expect such behavior. First, compact qualitative models of preference may (and typically do) admit more than one optimal (that is, non-dominated) solution. That, in principle, makes finding one such optimal solution easier. Second, if the preferences are captured by a TCP-net, then there are variable orderings ensuring that the first solution found will be an optimal one. In contrast, with GAI value functions, after we generate an optimal solution, typically we still have to explore the search tree to *prove* that no better solution exists. In the worst case, we have to explore the entire tree of CSPs, forcing us to explore a number of CSPs that is exponential in $|\mathcal{P}|$.

In summary, the first conclusion to be taken from our experiments is that subsets-space search fails to escape the trap of the large branching factor, while the stratified procedures for CSP-space search show a much higher potential. On the problems that require little backtracking in the space of CSPs, the latter procedures are actually very effective for both TCP-net and GAI function preference specification. Obviously, if the procedure is forced to explore many different CSPs, the performance unavoidably degrades. We note that, on larger databases, such backtracks often indicate an inherent conflict between desirable set-properties, and such conflicts might possibly be recognized and resolved off-line. In this work we do not investigate this issue, leaving it as an optional direction for future improvement.

The rather non-trivial example used in this section provides the reader also with the opportunity to assess the suitability of different preference specification languages. For example, although we used boolean-valued set properties, it may be argued that some of our natural-language preference statements would better be expressed using integer-valued set properties. Similarly, users may find that some other preference specification formalism, such as soft-constraints (Bistarelli et al., 1999), can more naturally capture these natural language preferences. This is an opportunity for us to reemphasize that while, for obvious reasons, we had to focus on a concrete choice of language, we believe that the two-tiered approach suggested here is far more general.

## 5. Complexity Analysis

Though reasonable runtimes have been obtained by us empirically with search over CSPs, both algorithm classes described above have a worst-case exponential running time. This begs the question of whether the problem itself is computationally hard. Obviously, with external constraints, subset optimization is NP-hard. Below we show that even without external constraints, the problem typically remains NP-hard, even with significant restrictions on the problem.

Naturally, the complexity of subset selection depends on the precise nature of the preference specification formalism used. Most of the results presented here assume TCP-net-based specification. Hardness results for this model immediately apply to the GAI model, based on an existing reduction (Brafman & Domshlak, 2008). In some cases, problems that are tractable under the TCP-net model become NP-hard when a GAI model is used, instead. Thus, unless stated otherwise, we assume henceforth that preferences over properties are specified by a TCP-net.

In analyzing the complexity of the problem we consider the following problem parameters:

- $n$, the overall number of items in the data set.

- $a$, the number of attributes of the items.

- $m$, the number of set properties, i.e. number of nodes in the TCP-net.

- $k$, maximal property formula size, defined as the number of logical connectives (and, or, not) in the formula.

- $d$ maximum attribute domain size, i.e. the maximum number of distinct values for each attribute.

- $\mu$, the number of times an attribute value can appear in the dataset.

### 5.1 NP-Hard Classes

**Theorem 1.** *When using TCP-based preferences over set properties, finding an optimal subset of a given set of items ($\mathcal{POS}$) is NP-hard even if the items are described only in terms of binary-valued attributes, and all the set properties are atomic (that is, we have $d = 2$ and $k = 0$).*

*Proof.* The proof is by a polynomial reduction from the well-known NP-hard Vertex Cover ($\mathcal{VC}$) problem. Given a graph $G = (V, E)$, a vertex cover of $G$ is a vertex subset $V' \subseteq V$ covering all the edges in the graph, that is, for every edge $e \in E$, there is a vertex $v \in V'$ such that $e$ is incident on $v$. The optimization version of $\mathcal{VC}$ corresponds to finding a minimal size vertex cover of $G$.

Given an $\mathcal{VC}$ problem instance $G = (V, E)$, we construct a $\mathcal{POS}$ problem instance by specifying a TCP-net $N$ and an item set $\mathcal{S}$ as follows. For each vertex $v \in V$ we create an item $o$ (denoted

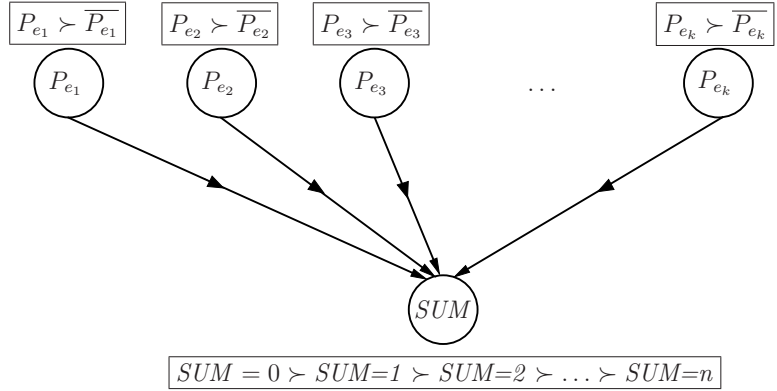

Figure 8: TCP-net in the reduction from $\mathcal{VC}$ to $\mathcal{POS}$ in the proof of Theorem 1.

by $o_v$), and thus we have $|\mathcal{S}| = |V| = n$ items. For each edge $e \in E$ we define an attribute $X$ (denoted by $X_e$), and thus we have $|\mathcal{X}| = |E| = a$ attributes. All the attributes in $\mathcal{X}$ are defined to have a binary, $\{0, 1\}$, domain. For each item $o_v$, the value of each attribute $X_e$ is $o_v[X_e] = 1$ if and only if $e$ is incident on $v$ in $G$. Next, for each edge $e \in E$, we define a binary set property $P_e = \langle |X_e| > 0 \rangle$ that takes the value $true$ if and only if at least one item in the selected subset provides the value 1 to the attribute $X_e$. In addition, we define a single multi-valued *empty* set property $SUM \equiv \langle || \rangle^5$. The domain of the *SUM* property is defined to be the integer-value range $[0..n]$. Note that, by construction, the properties utilize only one attribute per property, and thus no logical connectives, providing us with $k = 0$. The preferences over these set properties are

1. For each binary property $P_e$, the preference is for the value $true$, that is, $P_e \succ \overline{P_e}$.

2. For the empty property *SUM* we simply prefer smaller values, that is

$$(SUM = 0) \succ (SUM=1) \succ (SUM=2) \succ \ldots \succ (SUM=n)$$

The only edges in the TCP-net $N$, depicted in Figure 8, are the importance arcs from each $P_e$ to *SUM*, meaning that we would rather have to temporize in the value of the *SUM* property than have any of the $P_e$ being $false$.

Proposition 1 ensures that any optimal subset in the $\mathcal{POS}$ problem constructed as above always corresponds to a *proper* vertex cover of $G$.

**Proposition 1.** For any subset $S$ of $\mathcal{S}$ that is undominated with respect to the constructed TCP-net $N$, and every edge $e \in E$, we have $P_e(S) = true$.

*Proof.* Given an undominated (with respect to $N$) subset $S \subseteq \mathcal{S}$, let $P_e$ be a set property such that $P_e(S) = false$. By construction, there exists an item $o \in \mathcal{S}$ such that $o[X_e] = 1$. Considering $S' = S \cup \{o\}$, we have $S'$ being preferred to $S$ with respect to $N$ because (i) $S$ and $S'$ provide exactly the same values to all the set properties except for $P_e$ and *SUM*, (ii) $S$ provides a preferred

---

5. Since the formula $\varphi$ inside this set property is degenerate, and in fact equivalent to $\langle |true| \rangle$, every item in the selection set will have to comply with it. This set property is the simplest implementation of a counter

value to *SUM* while $S'$ provides a preferred value to $P_e$, and (iii) preferential improvement of $P_e$ dominates that of *SUM*. Thus $S'$ dominates $S$, contradicting the assumption that $S$ is undominated. □

**Lemma 1.** *For any subset $S$ of $\mathcal{S}$ that is undominated with respect to the constructed TCP-net $N$, there exists a vertex cover $V_S$ of $G$ with $|V_S| = |S|$.*

*Proof.* The proof is straightforward. Let $V_S = \{v \mid o_v \in S\}$. Because $S$ is undominated with respect to $N$, from Proposition 1 we have $P_e(S) = true$ for all binary "edge-related" properties $P_e$. In turn, $P_e(S) = true$ implies that $o[X_e] = 1$ for at least one item $o \in S$. By the construction, $o[X_e] = 1$ if and only if vertex $v$ covers edge $e$. Together with the mapping between the vertices $V$ and items $\mathcal{S}$ being bijective, the latter implies $|V_S| = |S|$. □

**Lemma 2.** *There exists a minimal vertex cover of $G$ of size $s$ if and only if there exists a subset $S \subseteq \mathcal{S}$ undominated with respect to $N$ such that $SUM(S) = s$.*

*Proof.* Let $S$ be an undominated subset of $\mathcal{S}$ with $|S| = s$. By construction, we have $P_e(S) = true$ for all binary set properties $P_e$, and $SUM(S') = s$. By Lemma 1, there exists a vertex cover $V_S$ of $G$ with $|V_S| = s$. Suppose to the contrary that $V_S$ is not minimal, that is, there exists a vertex cover $V'$ of $G$ with $|V'| < s$. Now, construct the subset $S' = \{o_v \mid v \in V'\}$. Since the mapping between $\mathcal{S}$ and $V$ is bijective, we have $|S'| = |V'| < s$, and thus $SUM(S') < s$. Likewise, by construction of our set properties and $V'$ being a vertex cover, we have $P_e(S) = true$ for all $P_e$. This, however, implies that $S'$ is preferred to $S$ with respect to $N$, contradicting the statement that $S$ is undominated. □

Theorem 1 now follows immediately from Lemma 2 and the fact that the reduction is clearly polynomial. □

**Theorem 2.** *Given TCP-based preferences over set properties, finding an optimal subset of a given set of items ($\mathcal{POS}$) is NP-hard even if the items are described in terms of a single attribute, all the set properties are binary-valued, each containing at most 2 logical connectives (that is, we have $a = 1$ and $k = 2$).*

*Proof.* The proof is by a polynomial reduction from $k$-SAT, for any $k \geq 3$. Given a $k$-SAT problem instance over propositional variables $V$ and logical formula $\Phi$, we construct a $\mathcal{POS}$ problem instance by specifying a TCP-net $N$ and an item set $\mathcal{S}$ as follows. For each variable $v \in V$, construct an item $o_v$ and an item $o_{\bar{v}}$, and thus $\mathcal{S}$ contains an item for every possible literal in the formula. The value of the only attribute $X$ is defined as follows: for each item $o_l$, we have $A(o_l) = l$ (where $l$ is a literal, either $v$ or $\bar{v}$, for all $v \in V$). The binary set properties $\mathcal{P}$ for the TCP-net $N$ are now defined as follows.

- Properties ensuring that a variable assignment is legitimate. For each variable $v \in V$,

$$P_v = \langle |X = v \vee X = \bar{v}| = 1 \rangle,$$

that is, for any $S \subseteq \mathcal{S}$, $P_v(S) = true$ if and only if $S$ contains exactly one of the items $\{o_v, o_{\bar{v}}\}$.

- Properties ensuring that $\Phi$ is satisfied. For each clause $C = (l_1 \vee l_2 \vee l_3 \vee ...) \in \Phi$:

$$P_C = \langle |X = l_1 \vee X = l_2 \vee X = l_3 \vee ...| \geq 1 \rangle$$

  that is, for any $S \subseteq \mathcal{S}$, $P_C(S) = true$ if and only if $S$ contains at least one item corresponding to a literal in $C$.

Finally, to complete the preference specification, we make all properties independent (that is, the TCP-net has *no edges*), and for each of the properties we prefer value *true* to value *false*.

To illustrate the above construction, consider a 3-SAT formula $\Phi = (x \vee \overline{y} \vee z) \wedge (y) \wedge (\overline{x} \vee z)$. For this formula, the construction leads to

| item | $X$ |
|------|-----|
| $o_x$ | $x$ |
| $o_{\overline{x}}$ | $\overline{x}$ |
| $o_y$ | $y$ |
| $o_{\overline{y}}$ | $\overline{y}$ |
| $o_z$ | $z$ |
| $o_{\overline{z}}$ | $\overline{z}$ |

**Set properties**:

$P_x = \langle |X = x \vee X = \overline{x}| = 1 \rangle$
$P_y = \langle |X = y \vee X = \overline{y}| = 1 \rangle$
$P_z = \langle |X = z \vee X = \overline{z}| = 1 \rangle$

$P_{C_1} = \langle |X = x \vee X = \overline{y} \vee X = z| \geq 1 \rangle$
$P_{C_2} = \langle |X = y| \geq 1 \rangle$
$P_{C_3} = \langle |X = \overline{x} \vee X = z| \geq 1 \rangle$

We now show that finding an undominated subset of $\mathcal{S}$ with respect to $N$ as above is equivalent to finding a satisfying assignment to $\Phi$. Let $S \subseteq \mathcal{S}$ be undominated with respect to $N$. We can show that $S$ provides value $true$ to all set propositions $P_v$ and $P_C$ (in this case we call $S$ an *ultimately preferred* subset) if and only if $\Phi$ is satisfiable.

First, let $S$ be an ultimately preferred subset of $\mathcal{S}$. Given such $S$, we can construct a mapping $\mathcal{A} : V \mapsto \{true, false\}$ such that $\mathcal{A}(v) = true$ if $o_v \in S$, and $\mathcal{A}(v) = false$ if $o_{\overline{v}} \in S$. Note that $\mathcal{A}$ is well-defined because, for an ultimately preferred subset $S$, all $P_v(S) = true$, and thus, for each $v \in V$, exactly one item from $\{o_v, o_{\overline{v}}\}$ is present in $S$. Clearly, $\mathcal{A}$ is a legal assignment for $\Phi$. In addition, we have all $P_C(S) = true$. Thus, for each clause $C \in \Phi$, at least one item with $X = l_i \in C$ belongs to $S$. By construction, this implies that $\mathcal{A}$ satisfies all the clauses in $\Phi$, and thus $\Phi$ is satisfiable.

Conversely, suppose that $S \subseteq \mathcal{S}$ is preferentially undominated with respect to $N$, but is not ultimately preferred. If our $\mathcal{POS}$ problem has such an undominated subset $S$, we show that $\Phi$ is unsatisfiable. Assuming the contrary, let $\mathcal{A}$ be a satisfying assignment of $\Phi$. Given $\mathcal{A}$, we construct a subset $S_\mathcal{A} \subseteq \mathcal{S}$ as $S_\mathcal{A} = \{o_l \mid \text{literal } l \in \mathcal{A}\}$, and show that $S_\mathcal{A}$ dominates $S$ with respect to $N$ (contradicting the assumed undominance of $S$, and finalizing the proof of Theorem 2).

By construction, since $\mathcal{A}$ is a legal assignment to $V$, we have $P_v(S_\mathcal{A}) = true$ for all set properties $P_v$. Also, since $\mathcal{A}$ is a satisfying assignment for $\Phi$, we have $P_C(S_\mathcal{A}) = true$ for all set properties $P_C$. Therefore, $S_\mathcal{A}$ is actually an ultimately preferred subset of $\mathcal{S}$. Finally, since all the set properties $\mathcal{P}$ are preferentially independent in $N$, and value $true$ is always preferred to value $false$ for all the set properties, we have that $S_\mathcal{A}$ dominates $S$ with respect to $N$. $\qquad \square$

Notice that Theorems 1 and 2 do not subsume each other. Theorem 1 poses no restriction on the number of item attributes in the problem instance, but does restrict the domain of all the attributes. Theorem 2 restricts the number of attributes to 1, but has no restriction on the domain size of this attribute, and its restriction on the property size is looser than that imposed in Theorem 1.

Finally, we note that tightening the condition of Theorem 2, by allowing only *at most 1 connective* in each set-property definition prevents us from using the same reduction as in the proof of Theorem 2 because the respectibe satisfiability problems would be the polynomial-time solvable 2-SAT problems. Our conjecture, however, is that this fragment of $\mathcal{POS}$ is still NP-hard. In fact, in Section 5.3 we show that the corresponding fragment of $\mathcal{POS}$ with the GAI preference specification (instead of TCP-nets) is indeed NP-hard.

## 5.2 Tractable Classes

Several tractable classes of $\mathcal{POS}$, obtained by further restricting the problem class discussed in Theorem 2, and characterized by single-attribute item description (that is, $a = 1$), are discussed below. In both trivially tractable (Section 5.2.1) and non-trivially tractable (Section 5.2.2) cases, we assume that the relational symbols are either equalities or inequalities, that in the specification of a property only equalities ("attribute = value") are used, and in addition we do not allow an empty set property to be specified. The latter restriction is due to the fact that the empty set property is somewhat special, as it enriches the descriptive power by allowing one to simulate an additional attribute in certain cases, and the single-attribute restriction is crucial for our tractability result.

Before we proceed with the actual results, note that, with a single-attribute item description, no two set properties can be in a conflict that demands backtracking while choosing items (i.e. during CSP solution). To illustrate such conflicts, consider the following examples.

1.
   1.a $\langle |A = a_i| \leq 5 \rangle$
   1.b $\langle |A = a_i| \leq 3 \rangle$    Set property 1.a is redundant, subsumed by 1.b

2.
   2.a $\langle |A = a_s| = 5 \rangle$
   2.b $\langle |A = a_s| > 6 \rangle$    One of these set properties must be false.

3.
   3.a $\langle |A = a_l| < 7 \rangle$
   3.b $\langle |A = a_l| \geq 9 \rangle$    One of these set properties must be false.

All such conflicts between set-properties can be resolved offline, prior to the actual process of subset selection, totally disregarding the available items. Hence, within the process of subset selection, we assume that there are no conflicts between set properties. Consequently, subset selection can be done in a greedy manner.

### 5.2.1 Trivially Tractable Class

**Theorem 3.** *Finding an optimal subset of a given set of items ($\mathcal{POS}$) with respect to a TCP-net preference specification is in P if the items are described in terms of a single attribute, and all the set properties are atomic (that is, we have $a = 1$ and $k = 0$).*

An algorithm for the problem class in Theorem 3 is depicted in Figure 9. The algorithm runs in time $\mathcal{O}(m^2 n)$, where $m$ is the number of set properties and $n$ is the number of available items $\mathcal{S}$. The **for** loop in line 4 of the algorithm iterates over all the set properties, each time checking compatibility with the previously considered properties, which requires $\Theta(m^2)$ time. The procedures GetSatisfyingSet($\cdot$) and HasSatisfyingSet($\cdot$) have to process each item in $\mathcal{S}$ only once. Hence, the total running time of the algorithm is $\mathcal{O}(m^2 n)$.[6]

---

6. This runtime analysis does not include the ordering of the TCP-net variables that is assumed to be given. One way to do that would be a topological sort of the net, that obviously can be done in polynomial time.

```
 1:  S_opt ← ∅
 2:  Fix a preference ordering over set properties 𝒫
 3:  𝒫_ass ← ∅
 4:  for each property P ∈ 𝒫 do
 5:      while (not (P.isSatisfied)) do
 6:          if P is in conflict with 𝒫_ass then
 7:              Set next value to P w.r.t. 𝒫_ass              ▷ Offline conflict resolution
 8:          else
 9:              if HasSatisfyingSet(P) then
10:                  S_opt ← S_opt ∪ GetSatisfyingSet(P)
11:                  P.isSatisfied ← true
12:                  𝒫_ass ← 𝒫_ass ∪ {P}
13:              end if
14:          end if
15:      end while
16:  end for
17:  return S_opt
```

```
 1:  procedure GetSatisfyingSet(P)
 2:      S ← ∅
 3:      for each item o ∈ 𝒮 do
 4:          if o has the property value defined by P then
 5:              S ← S ∪ {o}
 6:          end if
 7:          if |S| P.op P.cardinality then              ▷ If cardinality of S satisfies P
 8:              return S
 9:          end if
10:      end for
11:  end procedure
```

Figure 9: A polynomial-time algorithm for the $\mathcal{POS}$ problems with TCP-net preference specification, single-attribute item description, and all the set properties being atomic (that is, $a = 1$ and $k = 0$).

### 5.2.2 Non-Trivially Tractable Class

At the end of Section 5.1 we have mentioned that the complexity of $\mathcal{POS}$ under limiting the set-property description to at most one logical connective is still an open problem. If, however, we impose the limitations summarized in Table 4, we can show that the problem becomes tractable.

**Theorem 4.** *Finding an optimal subset of a given set of items ($\mathcal{POS}$) with respect to a TCP-net preference specification is in P if it is restricted as in Table 4.*

First we should discuss the implicit limitations (or special problem properties) that are imposed by the explicit limitations listed in Table 4.

---

1. All the items have only one attribute ($a = 1$)

2. All the property formulas have at most 1 connective ($k = 1$), and are positive (that is, we disallow negation)

3. The empty property is disallowed

4. The number of attribute value appearances is limited to at most $\mu = 1$ (that is, values in the attribute domain cannot be repeated)

---

Table 4: Characteristics of the tractable subclass of $\mathcal{POS}$ presented in Section 5.2.2.

1. The restriction to at most one attribute-value appearance in the data set provides a one-to-one correspondence between attribute values and items in $\mathcal{S}$. This means that each item can uniquely represent a specific attribute-value combination, and vice versa.

2. The restriction to a single-attribute item description renders the "$\wedge$" connective redundant. That is because the properties using the "$\wedge$" logical connective can only be of the form:

$$X = x_i \wedge X = x_j.$$

(Without loss of generality we assume $i \neq j$, or otherwise we can simply drop one of the terms.) These properties obviously cannot be satisfied because no item can have two different values for the only attribute $X$. In fact, set properties defined this way are equivalent to a property that is always $false$.

3. The only relevant cardinalities for the set properties are [0..2]. A property defined using only one connective with the restriction on the number of repetitions is not expressive enough to state a set property involving more than 2 items. If the *value* in a set property:

$$\langle |A = a_i \vee A = a_j| \quad op \quad value \rangle$$

is greater than 2, and $op \in \{\geq, >\}$, then again it cannot be satisfied. If the *op* of a property is $\leq$ or $<$, and the *value* is greater than 2, then it can be substituted by an effectively equivalent set property with *op* being $\leq$ and *value* = 2 .

The algorithm for the problem class in Theorem 4 is depicted in Figure 10. This algorithm bears some similarity to the algorithm in Figure 9, except that here the procedures GETSATISFYINGSET and HASSATISFYINGSET reason simultaneously about satisfaction of collections of set-property values, and do that by utilizing 2-SAT solving. Specifically, in Table 5 we show how any *valid* property in such a $\mathcal{POS}$ problem can be translated into a 2-SAT CNF formula. In Lemma 3 we prove the correctness of this translation. We should note that by using 2-SAT we can have an answer to the question "Is there a subset of items satisfying some already evaluated set-property values". The procedures GETSATISFYINGSET and HASSATISFYINGSET use the aforementioned reduction to 2-SAT to provide the answer in polynomial time.

**Lemma 3.** *There is a subset $S$ satisfying all the property-values $\mathcal{P}_{ass}$ if and only if there is a satisfying assignment $\mathcal{A}$ to the 2-SAT formula constructed from $\mathcal{P}_{ass}$.*

| | |
|---|---|
| $\langle |X = x_i| > 2 \rangle \Rightarrow$ infeasible | $\langle |X = x_i \vee X = x_j| > 2 \rangle \Rightarrow$ infeasible |
| $\langle |X = x_i| \geq 2 \rangle \Rightarrow$ infeasible | $\langle |X = x_i \vee X = x_j| \geq 2 \rangle \Rightarrow$ substituted by $\langle |X = x_i \vee X = x_j| = 2 \rangle$ and translated to $(v_i)$ and $(v_j)$ clauses |
| $\langle |X = x_i| \geq 1 \rangle \Rightarrow$ substituted by $\langle |X = x_i| = 1 \rangle$ and translated to $(v_i)$ clause | |
| | $\langle |X = x_i \vee X = x_j| \geq 1 \rangle \Rightarrow$ translated to $(v_i \vee v_j)$ clause |
| $\langle |X = x_i| \geq 0 \rangle \Rightarrow$ translated to $(v_i \vee \bar{v_i})$ clause | |
| | $\langle |X = x_i \vee X = x_j| = 1 \rangle \Rightarrow$ translated to $(v_i \vee v_j)$ and $(\bar{v_i} \vee \bar{v_j})$ clauses |
| $\langle |X = x_i| = 0 \rangle \Rightarrow$ translated to $(\bar{v_i})$ clause | |
| | $\langle |X = x_i \vee X = x_j| \geq 0 \rangle \Rightarrow$ translated to $(\bar{v_i} \vee \bar{v_j})$ clause |
| | $\langle |X = x_i \vee X = x_j| = 0 \rangle \Rightarrow$ translated to $(\bar{v_i})$ and $(\bar{v_j})$ clauses |

*Properties having **0** logical connectives*   *Properties having **1** logical connective*

Table 5: Translation of the set properties for the $\mathcal{POS}$ subclass in Section 5.2.2 to 2-SAT.

1:  Fix a preference ordering over set properties $\mathcal{P}$
2:  $S_{opt} \leftarrow \emptyset$
3:  $P_{ass} \leftarrow \emptyset$
4:  **for each** property $P \in \mathcal{P}$ **do**
5:      **while** (**not** ($P$.isSatisfied)) **do**
6:          Set next value to $P$ w.r.t. $\mathcal{P}_{ass}$
7:          **if** HASSATISFYINGSET($P_{ass}$) **then**                 ▷ Use reduction to 2-SAT
8:              $S_{opt} \leftarrow$ GETSATISFYINGSET($P_{ass}$)          ▷ Use reduction to 2-SAT
9:              $P$.isSatisfied $\leftarrow true$
10:             $P_{ass} \leftarrow P_{ass} \cup \{P\}$
11:         **end if**
12:     **end while**
13: **end for**
14: **return** $S_{opt}$

Figure 10: A poly-time algorithm for the $\mathcal{POS}$ problems with TCP-net preference specification, and characteristics as in Table 4.

*Proof.* By construction, we have an injective correspondence between the properties in the $\mathcal{POS}$ problem and clauses in the 2-SAT problem. Every property $P \in \mathcal{P}$ injectively corresponds to a certain clause $\varphi_P$. Every item $o \in \mathcal{S}$ injectively corresponds to a propositional variable $v_i \in V$. Thus, the correspondence between the selected subset $S$ and the assignment $\mathcal{A}$ is simply

$$v_i = true \quad \Leftrightarrow \quad o \in S. \tag{1}$$

Because the translation is injective and rather straightforward (without introducing any auxiliary clauses or properties), it is trivial that $S$ is a subset that satisfies all the properties in $\mathcal{P}_{ass}$ if and only if $\mathcal{A}$ is an assignment that satisfies all the clauses in the corresponding 2-SAT formula. □

The above shows correctness of the algorithm in Figure 10, and finalizes the proof of Theorem 4.

### 5.3 Complexity of $\mathcal{POS}$: TCP-nets vs. GAI Preference Specification

With the restrictions as in Table 4 we were able to show that the $\mathcal{POS}$ problem with TCP-net preference specification is tractable by reduction to 2-SAT, because there is no need to backtrack while searching in the attribute value space. An interesting question is, what if the specification were done using GAI functions?

**Theorem 5.** *Finding an optimal subset of a given set of items ($\mathcal{POS}$) with respect to a GAI preference specification is NP-hard even if the items are described in terms of a single attribute, all the set properties are binary-valued, each containing at most 1 logical connective (that is, we have $a = 1$ and $k = 1$).*

*Proof.* The proof is by a polynomial reduction from MAX-2SAT. As far as item definitions and properties are concerned, the reduction is essentially the same as the reduction from $k$-SAT in the proof of Theorem 2. That is, for each variable $v \in V$, construct an item $o_v$ and an item $o_{\bar{v}}$. The value of the only attribute $X$ is defined as follows: for item $o_l$, we have $A(o_l) = l$ (where $l$ is a literal, either $v$ or $\bar{v}$, for all $v \in V$). Set properties are also as in the proof of Theorem 2, but now they are limited to only 2 variables per clauses (re-stated for convenience below):

- For each variable $v \in V$:

$$P_v = \langle |X = v \vee X = \bar{v}| = 1 \rangle,$$

  that is, properties ensuring that a variable assignment is legitimate.

- For each clause $C = (l_1 \vee l_2) \in \Phi$:

$$P_C = \langle |X = l_1 \vee X = l_2| \geq 1 \rangle,$$

  that is, properties ensuring that $\Phi$ is satisfied.

The value function specification is such that legitimate variable assignments are enforced, and a larger number of clauses satisfied is preferred. This is achieved by using an additively independent value function (i.e., where each factor contains a single variable), with values being as follows. Each clause-satisfying property has a value of 1 for being $true$, and 0 for being $false$. Each literal-satisfying property has a value of 0 for being $true$, and a negative value of $-2m$ for being $false$, where $m$ is the number of clauses.

**Lemma 4.** *Given a GAI value function and item set $\mathcal{S}$ constructed as above for a 2-CNF formula $\Phi$, there exists a subset $S \subseteq \mathcal{S}$ with value of $U(S) = p$ if and only if there exists an assignment $\mathcal{A}$ satisfying $p$ clauses in $\Phi$.*

*Proof.* Let $S$ be any subset of $\mathcal{S}$ that has non-negative value. This implies by construction (since there are only $m$ clause-satisfying properties $P_C$) that all literal-satisfying properties must be true for $S$, and the respective assignment $\mathcal{A}_S$ can be constructed as in Equation 1. Conversely, let $\mathcal{A}$ be a legitimate assignment to the variables $V$. One can define a corresponding set $S_{\mathcal{A}}$, for which (by construction) all properties $P_v$ are $true$. Also, observe that by construction the number of $P_C$ set properties that are true on $S_{\mathcal{A}}$ is the same as the number of clauses satisfied by the assignment $\mathcal{A}$. □

The theorem follows immediately from the properties of the construction of the set properties and preferences. □

At the end of Section 5.1 we have noted that if the restrictions on the problem parameters are more severe than in Theorem 2, by limiting the number of logical connectives per set property to at most 1, we can no longer show whether the problem is tractable or NP-hard under the TCP-net preference specification. However, Theorem 5 shows that, with preferences specified using a GAI value function, the problem is in fact NP-hard. Moreover, the problem class from Theorem 5 subsumes the class from Theorem 4, and thus provides an additional result showing that even though with the TCP-net specification the respective problem is tractable, with a GAI preference specification it becomes NP-hard.

## 6. Related Work

In the introduction, we mentioned the closely related work of desJardins and Wagstaff (2005). In that approach, the motivation to provide the user with a diverse collection of values is either to reflect the set of possible choices better for applications where the user must eventually select a single item, or when the diversity of the selected set is an objective on its own. The work of Price and Messinger (2005) is explicitly concerned with this problem. Specifically, they consider the problem of recommending items to a user, and view it as a type of subset selection problem. For example, suppose we want to recommend a digital camera to a user. We have a large set of available cameras, and we are able to recommend $k$ cameras. Price and Messinger consider the question of how to select this set, proposing that the candidate set will maximize the expected value of the user's choice from this set. They suggest a concrete algorithmic approach for handling this problem. The input to their problem is some form of partial representation of the user's preferences (which can be diverse, as in our work) and naturally, the concrete techniques are different from ours. Both these papers share the assumption on ranking sets, common to most previous work as discussed by Barberà et al. (2004), that ultimately one item will be selected from this set. However, they do not necessarily start out with an initial ranking over single items, and as in our case, the work of desJardins and Wagstaff utilizes the attribute value of items in the selection process.

Earlier work on ranking subsets was motivated by problems such as the college admissions problem (Gale & Shapley, 1962), where we need to select the best set of fixed cardinality among a pool of college candidates. The admissions officer has various criteria for a good class of students and wishes to come up with an optimal choice. Some of the key questions that concerned this line of work were what are good properties of such set rankings and whether they have some simple representation. An example of a property of the set ranking that may be desirable is the following: given a set $S$, if we replace some member $c \in S$ with some other member $c'$ to obtain the set $S'$, and $c'$ is preferred to $c$, then $S'$ is preferred to $S$. An example of a representation of the ranking is an additive representation where items are associated with real values and one set is preferred to another if the sum of its elements' values is larger. It would be interesting to study similar question in our context of structured objects.

This question of ranking sets appears in other areas, such as logics of preference and likelihood. For example, the main question considered by Halpern (1997) is how to construct an ordering over formulas based on an ordering over truth assignments. Formulas are associated with the set of worlds in which they are satisfied, and hence, the question of comparing the likelihood of formulas $\psi$ and $\phi$ corresponds to that of ranking their respective set of models given the initial ranking on single models. Much work on non-monotonic logics uses Shoham's preference semantics (Shoham, 1987), and semantically, such work (see, e.g., Kraus, Lehmann, & Magidor, 1990) can be viewed as attempting to answer the opposite question – define a ranking over single truth assignments given some, possibly partial, ordering over formulas, i.e., sets of models.

A number of lines of work are related to our specification and solution methods. The first is the work on Russian Doll Search (RDS), a well known algorithm for combinatorial optimization, originally presented by Verfaillie, Lemaître, and Schiex (1996) as an efficient algorithm for Constraint Optimization Problems (COP). The idea behind the approach is to solve consecutively harder problems. Initially, the problem is solved while considering only one variable. The optimal result provides a lower bound. Each iteration, additional variables are considered, until eventually the original problem is solved. By using the lower bound obtained from the previous iteration (and other optimizations) this technique is often able to solve the original problem more efficiently. Recently Rollon and Larrosa (2007) extended Russian Doll Search to support multi-objective optimization problems. In a multi-objective optimization problem the goal is to optimize several parameters (attributes) of the variables in the problem. Usually all the parameters cannot be simultaneously optimized. The technique of Rollon and Larrosa involves incremental solution with more and more objectives included, and, in this sense, it is related to our search over CSPs approach in which we incrementally consider more and more set properties. Indeed, different desirable set properties can be viewed as different objectives.

Another related area is that of Pseudo-Boolean Constraint (PBC) Satisfaction Problems (Sheini & Sakallah, 2005). A PBC has the form:

$$\sum_i w_i l_i \geq k.$$

Here the $l_i$'s are literals and we interpret their values as being either 0 (*false*) or 1 (*true*); the $w_i$ are real-valued coefficients; and $k$ is an integer. Thus Pseudo-Boolean CSPs are a special form of integer programs, and can nicely represent the cardinality constraints we generate. Thus, one option for solving the type of CSPs generated here would be using a dedicated PBC solver. We run several popular PBC solvers on the satisfiability instances generated during the optimization: *Pueblo* (Sheini & Sakallah, 2005), *MiniSat* (Eén & Sörensson, 2005), and *Galena* (Dixon & Ginsberg, 2002). These solvers showed comparable results for satisfiable cases, while for the unsatisfiable cases, the PBC solvers showed better performance. This appears to be due to their use of linear programming as a preliminary test for satisfiability.

Another line of work that bears important connection to ours is that of winner determination in combinatorial auctions. In regular auctions, bidders bid for a single item. In combinatorial auctions, bidders bid on bundles of items. Thus, bidders must provide their preferences over different subsets of the set of auctioned items. The goal in combinatorial auctions is to allocate the set of goods to different bidders in the best manner (e.g., maximizing the payment to the seller or maximizing total welfare). This differs from the problem of selecting a single optimal subset with which we are concerned. However, in both cases, preferences over subsets must be provided to the optimization algorithm. As the number of subsets is exponential in the number of items, researchers in combinatorial auctions have sought bidding languages that can succinctly describe preferences of interest (Boutilier & Hoos, 2001; Nisan, 2006). What distinguishes our specification approach is its reliance on the existence of item features and the desire to provide a generic specification that does not depend on the concrete set of items. Work in combinatorial auctions also attempts to break the specification in some way. This is typically done by specifying values for small bundles and providing rules for deriving the value of larger sets from the values of the smaller sets.

## 7. Conclusion

We suggested a simple, yet general approach to lifting any attribute-based preference specification formalism to one for specifying preferences over sets. We then focused one instantiation of this idea via a concrete language for specifying set properties, and suggested two methods for computing an optimal subset given such a specification. One method is based on searching the space of explicit subsets, while the other searches over implicit subsets represented as CSPs. Both search spaces are meaningful regardless of the specific underlying preference specification algorithm although the precise search and bounds generation method will vary. We focused on two concrete and popular specification formalisms, one qualitative and one quantitative, on which we experiment and provide complexity results. Although the problem is generally NP-hard, as expected, the experimental results are quite encouraging.

We wish to reemphasize that other choices, both for the set property language and the preference specification formalism are possible, and may be more appropriate in various cases. Indeed, an interesting topic for future research would be to see which choices fit best some natural application areas; whether and how the algorithm presented in this paper can be modified to handle such languages; and how the complexity of the optimal subset selection problem is affected by such choices.

Though incremental search over CSPs appears to be the better method for optimal subset selection, it leaves a few questions open. First, it is an interesting question whether an efficient NoGood recording scheme that does not rely on static variable and value orderings exists. Intuitively, such a scheme should exist since the CSPs generated can be efficiently encoded into SAT as a boolean CNF formula (Bailleux & Boufkhad, 2004; Eén & Sörensson, 2005), and clause learning is a well known technique in SAT solving. Second, we have seen that while the incremental approach usually improves the overall performance, its contribution is not substantial and what really improves the performance is better individual CSP solving. This begs two questions: (1) Can we better utilize solutions across CSPs, and (2) Would representing and solving the CSPs generated as pseudo-boolean CSPs (Manquinho & Roussel, 2006) or SAT instances lead to faster solution times? Naturally, alternative approaches are also feasible.

Finally, in various applications, the set of elements gradually changes, and we need to adapt the selected subset to these changes. An example is when we use this approach to choose the most interesting current articles, and new articles constantly appear. It is likely that in this case the preferred set is similar to the current set, and we would like to formulate an incremental approach that adapts to such changes quickly.

## Acknowledgments

Preliminary versions of this work appeared in (Brafman, Domshlak, Shimony, & Silver, 2006b; Binshtok, Brafman, Shimony, Mani, & Boutilier, 2007). The authors wish to thank our anonymous reviewers for their useful comments and suggestions. Brafman was supported in part by NSF grant IIS-0534662, Brafman and Domshlak were supported by the COST action IC0602, Binshtok, Brafman and Shimony were supported by Deutsche Telekom Laboratories at Ben-Gurion University, by the Paul Ivanier Center for Robotics Research and Production Management, and by the Lynn and William Frankel Center for Computer Science.